# DKGCM: A Spatio-Temporal Prediction Model for Traffic Flow by Fusing Spatial Node Clustering Method and Fourier Bidirectional Mamba Mechanism

## Author Information

Siqing Long[1], Xiangzhi Huang[2], Jiemin Xie[3], Ming Cai[4] (*Corresponding author*).

1. PhD Candidates, Guangdong Provincial Key Laboratory of Intelligent Transportation System, School of Intelligent Systems Engineering, Shenzhen Campus of Sun Yat-sen University, Shenzhen 518107, China; email: longsq@mail2.sysu.edu.cn

2. PhD Candidates, Guangdong Provincial Key Laboratory of Intelligent Transportation System, School of Intelligent Systems Engineering, Shenzhen Campus of Sun Yat-sen University, Shenzhen 518107, China; email: huangxzh33@mail2.sysu.edu.cn

3. Associate Professor, Guangdong Provincial Key Laboratory of Intelligent Transportation System, School of Intelligent Systems Engineering, Shenzhen Campus of Sun Yat-sen University, Shenzhen 518107, China; email: xiejm28@mail.sysu.edu.cn

4. Professor, Guangdong Provincial Key Laboratory of Intelligent Transportation System, School of Intelligent Systems Engineering, Shenzhen Campus of Sun Yat-sen University, Shenzhen 518107, China; email: caiming@mail.sysu.edu.cn (Corresponding author)

# DKGCM: A Spatio-Temporal Prediction Model for Traffic Flow by Fusing Spatial Node Clustering Method and Fourier Bidirectional Mamba Mechanism

**Abstract:**

Accurate traffic demand forecasting enables transportation management departments to allocate resources more effectively, thereby improving their utilization efficiency. However, complex spatiotemporal relationships in traffic systems continue to limit the performance of demand forecasting models. To improve the accuracy of spatiotemporal traffic demand prediction, we propose a new graph convolutional network structure called DKGCM. Specifically, we first consider the spatial flow distribution of different traffic nodes and propose a novel temporal similarity-based clustering graph convolution method, DK-GCN. This method utilizes Dynamic Time Warping (DTW) and K-means clustering to group traffic nodes and more effectively capture spatial dependencies. On the temporal scale, we integrate the Fast Fourier Transform (FFT) within the bidirectional Mamba deep learning framework to capture temporal dependencies in traffic demand. To further optimize model training, we incorporate the GRPO reinforcement learning strategy to enhance the loss function feedback mechanism. Extensive experiments demonstrate that our model outperforms several advanced methods and achieves strong results on three public datasets.

## 1. Introduction

The rapid growth of travel demand and the continuous increase in the number of automobiles make traffic congestion an increasingly severe problem in global cities. To address this issue, proactive traffic management strategies based on predicted values of traffic flow (such as traffic volume, speed, and density) are needed. For instance, strategies such as ramp metering and traffic signal optimization can be adopted to improve road utilization and enhance citizens' travel efficiency. However, to design such solutions, a prerequisite is accurate spatiotemporal traffic flow prediction, i.e., forecasting future traffic flow based on historical data collected from traffic sensors spatially distributed across a road network.

Various methods have been proposed for spatiotemporal traffic flow prediction, and the most commonly used approaches in the past include statistical-based methods, machine learning based methods, and Convolutional Neural Network(CNN) based methods (See Section 2 for a review of related studies). Among these, one of the most sophisticated methods is those based on Graph Neural Network (GNN) (Wang et al., 2020). GNNs model road networks by representing non-Euclidean spatial dependencies as a graph and can aggregate and propagate relevant traffic information. There are usually two components in these methods, one responsible for learning the spatial dependencies (such as GCN or GAT) and the other for learning the temporal dynamics (such as LSTM or Transformer).

The association graph defines how traffic information propagates spatially; the accuracy of this graph largely determines the performance of GNNs, and thus, constructing an accurate association graph is critical for spatiotemporal traffic flow prediction. Broadly speaking, two types of association graphs have been used in existing studies: static graphs and dynamic graphs. Static graphs rely on fixed prior knowledge to define node relationships, such as road topology or historical traffic patterns. For example, T-GCN utilizes a predefined graph structure to model node correlations (Zhao et al., 2020), while STFGNN employs dynamic time warping to measure node similarity (M. Li & Zhu, 2021). Dynamic graphs adaptively infer node relationships from data, enabling the model to capture evolving spatial dependencies. AGCRN, for instance, uses node embeddings to infer dynamic relationships between nodes (Bai et al., 2020). DSTAGNN leverages the Wasserstein distance to evaluate node similarity and build dynamic graphs accordingly (Lan et al., 2022). Graph Wave Net introduces learnable node representations, optimizing them via stochastic gradient descent to generate an adaptive adjacency matrix (Wu et al., 2019). In addition, some studies have chosen to incorporate external influencing factors (such as the number of lanes and road type) to construct adaptive dynamic adjacency matrices (Y. Xu et al., 2023), thereby enhancing the interpretability of spatial feature extraction processes (Bao et al., 2023).

However, both static and dynamic graphs exhibit certain limitations. Static graph-based methods are unable to capture structural dynamics. Their fixed graph structures fail to reflect evolving connection relationships or significantly different interaction patterns across time periods. As a result, these methods lack sensitivity to temporal feature variations and are limited in modeling time-dependent changes in the data. As illustrated in Figure 1, although Nodes 1 and 2 are spatially adjacent, they demonstrate substantial differences in both the magnitude and distribution of traffic flow over time. Conversely, Nodes 1 and 16, while geographically distant, exhibit remarkably similar traffic flow dynamics. This discrepancy underscores the necessity of developing more robust approaches that transcend spatial proximity and incorporate temporal feature similarity as a complementary dimension for capturing spatiotemporal correlations.

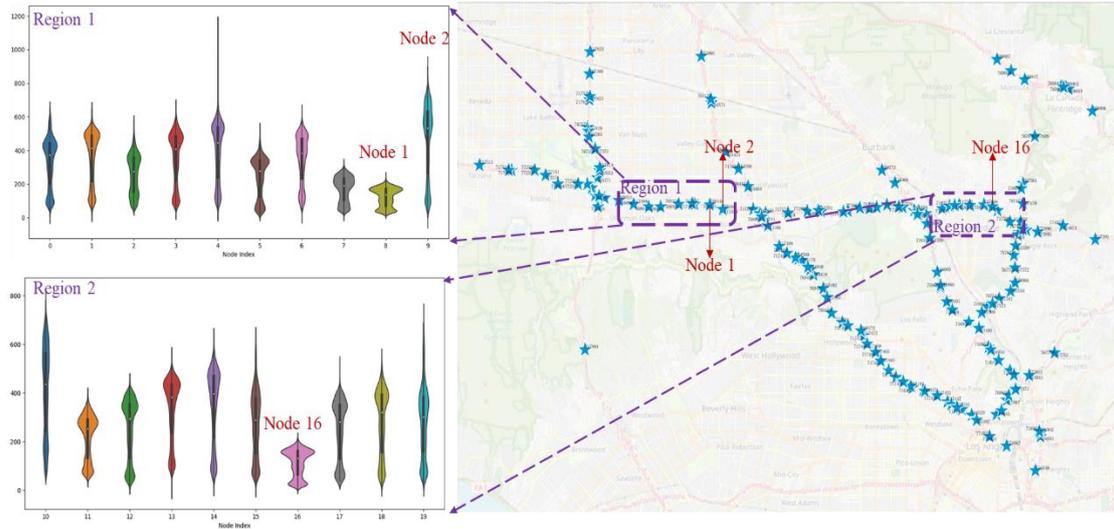

Figure 1. Spatial and temporal heterogeneity of traffic flow between different detection loops

On the contrary, Dynamic graphs can adapt to real-time traffic dynamics (such as sudden congestion), thereby reducing reliance on imperfect prior knowledge. However, these methods often incur high computational costs and carry the risk of overfitting if inappropriate regularization is applied. Meanwhile, many dynamic graph models are highly sensitive to short-term and instantaneous fluctuations in graph structure, often overlooking the underlying evolution patterns that emerge over more stable and macroscopic time scales. Therefore, we propose a clustering-based graph convolutional method, DK-GCN, which is grounded in node temporal clustering. The core idea of this approach is to segment the entire spatial graph into several homogeneous sub-regions based on temporal similarity among nodes. A static subgraph is then constructed within each sub-region, where a spatiotemporal graph convolutional network is applied. This method aims to strike a balance between dynamic adaptability and computational efficiency, while effectively capturing stable temporal patterns.

In terms of capturing temporal features, most existing methods rely on structures such as CNN, Recurrent Neural Networks (RNN), including Long Short-Term Memory (LSTM) networks, and Transformer-based models. For example, DCRNN employs

Gated Recurrent Units (GRU) to capture temporal dependencies (Y. Li et al., 2018), while GMAN utilizes attention mechanisms for time series modeling (C. Zheng et al., 2020). However, RNN-based models often suffer from issues such as parameter forgetting and gradient vanishing when dealing with long-range dependency extraction tasks. In contrast, although Transformer-based models capture long-range dependencies more effectively, they are associated with high computational complexity. Therefore, there remains a lack of more advanced and effective approaches for modeling complex temporal correlations.

To solve this problem, the Mamba architecture introduces a novel state-space model (SSM) that effectively mitigates the vanishing gradient problem commonly associated with RNNs and eliminates the quadratic computational complexity inherent in Transformers. However, Mamba's core computation relies on recurrent scanning, whereby the contextual information of previous tokens is only accessible when processing the current token. This results in a unidirectional limitation and a lack of complementary mechanisms for understanding and representing time series from alternative perspectives—such as the frequency domain. To address these limitations, we propose the FBMamba architecture. This model aims to integrate the global representation capabilities of the frequency domain with the fine-grained local representation strength of the time domain, thereby generating more robust sequence representations. Additionally, FBMamba incorporates a bidirectional Mamba mechanism to overcome the limitations of one-way scanning and alleviate the information loss problem inherent in unidirectional modeling.

In addition, although GNN-based deep learning models have achieved remarkable progress in spatiotemporal traffic forecasting, their training pipelines remain suboptimal due to two fundamental limitations. First, relatively few studies have explored forward propagation, loss calculation, and backpropagation mechanisms in depth. As a result, these models are prone to issues such as overfitting or underfitting during training, ultimately leading to inefficient training and limited generalization capability. Second, the prevalent use of static loss functions during training, such as MAE and MSE, poses a challenge. These traditional loss functions serve as fixed optimization objectives throughout the entire training process and cannot adapt to the model's evolving convergence needs—for instance, prioritizing the capture of coarse-grained patterns in the early stages and focusing on minimizing fine-grained errors in later stages.

Therefore, motivated by the aforementioned challenges, this paper proposes a novel spatiotemporal graph clustering convolutional network (DKGCM) to achieve more accurate traffic flow prediction. The key contributions of this work are as follows:

- ✓ **New spatial scale modeling approach**: A clustering-based graph convolutional framework, DK-GCN, is introduced. This method leverages Dynamic Time Warping (DTW) to extract time-series features of traffic nodes and applies K-means clustering to group similar nodes. Subsequently, Graph Convolutional Networks (GCN) are employed to capture spatial dependencies within each cluster.
- ✓ **Effective temporal scale modeling methods**: The model incorporates the Fast Fourier Transform (FFT) algorithm with the bidirectional Mamba time series prediction framework, enabling efficient and accurate modeling of temporal dependencies with reduced computational complexity.
- ✓ **Optimized model training process**: To enhance training stability and mitigate overfitting/underfitting issues, a reinforcement learning strategy, Gradient Reinforcement Policy Optimization (GRPO), is integrated into the loss function optimization process.
- ✓ **Comprehensive experimental validation**: Extensive experiments conducted on three publicly available traffic datasets—PEMS04, PEMS07, and PEMS08—demonstrate the superior performance of the proposed DKGCM model. Results show significant improvements in prediction accuracy, with notable reductions in Mean Absolute Error (MAE) and Root Mean Square Error (RMSE) compared to baselines.

The remainder of this paper is organized as follows. Section 2 provides a comprehensive review of existing methods for spatiotemporal traffic flow prediction. Section 3 presents the preliminary concepts and foundational knowledge relevant to this study. Section 4 details the structure and methodological components of the proposed DKGCM model. Section 5 conducts extensive experiments to evaluate the model's performance and offers an in-depth analysis and discussion of the results. Finally, Section 6 concludes the paper by summarizing the key contributions and findings and outlining potential directions for future research.

## 2. Related Work

The task of spatiotemporal traffic flow prediction has currently led to many solutions. Traditional methods, including ARIMA (Van Der Voort et al., 1996), Kalman filters (van Hinsbergen et al., 2012), and support vector machines (SVM) (Castro-Neto et al., 2009), have been widely explored, but their performance is limited. The advancement of deep learning methods allows better modeling of the complex nonlinear relationships within traffic patterns on urban road networks and has quickly become the dominant method for spatiotemporal traffic flow prediction.

Early studies in this direction adopted CNNs and RNNs to process spatial and temporal information, respectively (Ma et al., 2017) (G. Zheng et al., 2023), which laid the foundation for subsequent research. However, CNNs are typically designed to process regular grid data, such as images, whereas traffic networks exhibit graph structures and non-Euclidean properties, making them difficult for CNNs to model effectively. Thus, GNNs have been adopted to address this challenge(D. Cao et al., 2020). For example, STGCN utilizes one-dimensional CNN and GCN to capture spatiotemporal dependencies (B. Yu et al., 2018). DCRNN incorporates diffusion models to simulate the traffic flow diffusion process and enhance GRU structures. ASTGCN models three distinct periods to improve the model's capture of time dependencies (Guo et al., 2019). These models have significantly improved the accuracy of spatiotemporal traffic flow prediction. However, none of the above methods specifically focus on optimizing the construction of graph relationships.

To address this challenge, researchers are now moving beyond traditional neural network combinations like CNN-LSTM (M. Cao et al., 2020), GCN-LSTM (Z. Li et al., 2019), and GCN-GRU (Tao et al., 2020). Some studies focus on optimizing traditional predefined graph structures (Song et al., 2020) and propose self-learning graph structures to mitigate biases in feature acquisition. Models like MTGNN (Wu et al., 2020) and STEP (Shao, Zhang, Wang, & Xu, 2022) are examples of this approach. In addition, the combination of the adaptive adjacency matrix and node embeddings in Graph WaveNet offers a new perspective for graph modeling. Other studies have moved away from graph structures entirely, opting for non-graph models for spatiotemporal prediction. For instance, STNorm (Deng et al., 2021) uses spatiotemporal regularization to simplify the model, while STID (Shao, Zhang, Wang, Wei, et al., 2022) introduces a spatiotemporal identity addition method to reduce computational complexity. Despite their simplicity, these models also demonstrate strong prediction performance.

In addition to graph structures, the study of temporal dependency extraction modules is equally important. Traditional methods primarily use RNNs and their derivatives, such as LSTM (Cui et al., 2020) and GRU, for temporal feature extraction.

However, RNN-based approaches often suffer from issues like error accumulation and slow training speeds. In recent years, the introduction of Transformer models and the widespread application of Graph Attention Networks in spatiotemporal prediction (M. Xu et al., 2021) have proven that attention mechanisms can significantly improve model performance (Park et al., 2020). Furthermore, models like Informer (Zhou et al., 2021), which employs a multi-head probabilistic sparse attention mechanism, and its variants, such as Crossformer (Y. Zhang & Yan, 2022) using cross-attention mechanisms and iTransformer (Y. Liu et al., 2023) with an inverted attention structure, have shown effectiveness in capturing temporal dependencies within sequences. Despite these advancements, challenges like high time complexity and resource consumption persist. As a result, structures like Mamba (Cai et al., 2024), which introduce state-space vectors to reduce complexity and enhance stability, have gained popularity. Additionally, the MGCN model (Lin et al., 2025), which integrates GCN and Mamba, has demonstrated excellent performance in long-sequence prediction tasks.

The structural composition and main usage methods of the above spatiotemporal prediction model are summarized in Table 1.

Table 1 Overview of spatiotemporal prediction methods

| Model | Spatial topology construction | Spatial dependency | Temporal dependency | External method |
|---|---|---|---|---|
| DCRNN | Distance graph | GCN | GRU | None |
| TGC-LSTM | Binary graph | GCN | LSTM | None |
| AGCRN | Adaptive graph | GCN | GRU | None |
| STGNN(Wang et al., 2020) | Adaptive graph | GCN | GRU+Transfomer | None |
| GMAN | Distance graph | GAT | Embedding+Attention | None |
| ST-GRAT | Distance graph | GAT | Embedding+Attention | None |
| STTNs | Distance graph | GAT | Transfomer | None |
| STGCN | Distance graph | GCN | CNN | None |
| ASTGCN | Binary graph | GCN | CNN+Attention | None |
| STSGCN | Binary graph | GCN | GCN | None |
| GWNet | Adaptive graph | GCN | CNN | None |
| MTGNN | Adaptive graph | GCN | CNN | None |
| STNorm | None | Embedding | Embedding+MLP | None |
| STID | None | Embedding | MLP | None |
| STAEFormer(H. Liu et al., 2023) | None | Transformer | Transformer | Embedding |
| STD-MAE(Gao et al., 2024) | None | Patch+Trans | Patch+Trans | Masked Attention |
| MGCN | Distance graph | GCN | Mamba | None |
| Ours | Distance graph | DK-GCN | FFT+Mamba | GRPO RL |

As shown in Table 1, improving the effectiveness of traffic flow prediction models can be achieved from four key perspectives: spatial topology, spatial feature extraction methods, temporal feature extraction methods, and additional techniques. This paper introduces a novel traffic demand prediction model that focuses on enhancing spatiotemporal feature extraction through clustering, GCN, and the Mamba method. Additionally, the model incorporates the GRPO reinforcement learning strategy to optimize the training process, thereby improving overall model performance.

## 3. Preliminaries

**Definition 1**: In this study, we define a road traffic sensor network as an undirected, connected graph $\mathcal{G} = (\mathcal{V}, \mathcal{A})$, where $\mathcal{V}$ is the set of nodes (i.e., locations where traffic sensors are deployed) and $\mathcal{A}$ is the set of links between nodes, describing whether traffic sensors are connected by a route in the physical road network. For the convenience of notation, we define $N=|\mathcal{V}|$ as the number of nodes in the network. With this, link set $\mathcal{A}$ can be defined as an adjacency matrix $\mathbb{R}^{N \times N}$ that describes the topological relationships between the roads.

**Definition 2:** Let the traffic flow data at all nodes in the traffic sensor network $\mathcal{G}$ at a certain time $t$ be denoted as $x_t \in \mathbb{R}^N$, The historical traffic flow data over the past $H$ time steps for all $N$ nodes is represented as $X = [x_{t-H+1}, \cdots\cdots, x_t] \in \mathbb{R}^{H \times N}$, The goal is to predict the future traffic flow data $y = [x_{t+1}, \cdots\cdots, x_{t+F}] \in \mathbb{R}^{F \times N}$. That is, we aim to establish a prediction model with the mapping relationship $y = f(X, \mathcal{A})$, where $\mathcal{A}$ is the adjacency matrix of the traffic sensor network.

**Definition 3:** The SSM method is the core structure of the Mamba model, which is used to describe the inherent hidden states in the network. It can be incorporated into an end-to-end neural network architecture. The SSM primarily utilizes a set of first-order differential equations to track the evolution of network states over time. Specifically, given the input traffic flow history time series $x_t \in \mathbb{R}^N$, the model transforms it into the output future traffic flow time series $y_t \in \mathbb{R}^N$ through the potential state vector $h_t \in \mathbb{R}^N$, as shown in Formula 1.

$$\begin{aligned} h_t' &= A \cdot h_t + B \cdot x_t \\ y_t &= C \cdot h_t \end{aligned} \quad (1)$$

In the State Space Model, $A, B, C \in \mathbb{R}^{N \times N}$ are learnable matrices, The continuous parameters $(\Delta, A, B)$ can be discretized into discrete parameters $(\bar{A}, \bar{B})$, where $\bar{A} =$

$\exp(\Delta, A)$, and $\bar{B} = (\Delta A^{-1}) \cdot (\exp(\Delta A) - I) \cdot \Delta B$. the calculation process can be carried out through either linear recursion or global convolution (Gu & Dao, 2024).

## 4. Proposed Model and Methodology

The structure of the DKGCM spatiotemporal traffic flow prediction model, as proposed in this paper, is shown in Figure 1. The model consists of three main components: the DK-GCN spatial extraction module, the FBMamba temporal extraction module, and the output module. The DK-GCN module clusters traffic nodes with high feature similarity and extracts spatial features using a layer of spatiotemporal graph convolutional network. The FBMamba module, incorporates the Fast Fourier Transform (FFT) algorithm with the bidirectional Mamba mechanism to extract temporal features for the corresponding node category. The output module uses a fully connected layer to generate hidden layer parameters and GRPO strategy vectors.

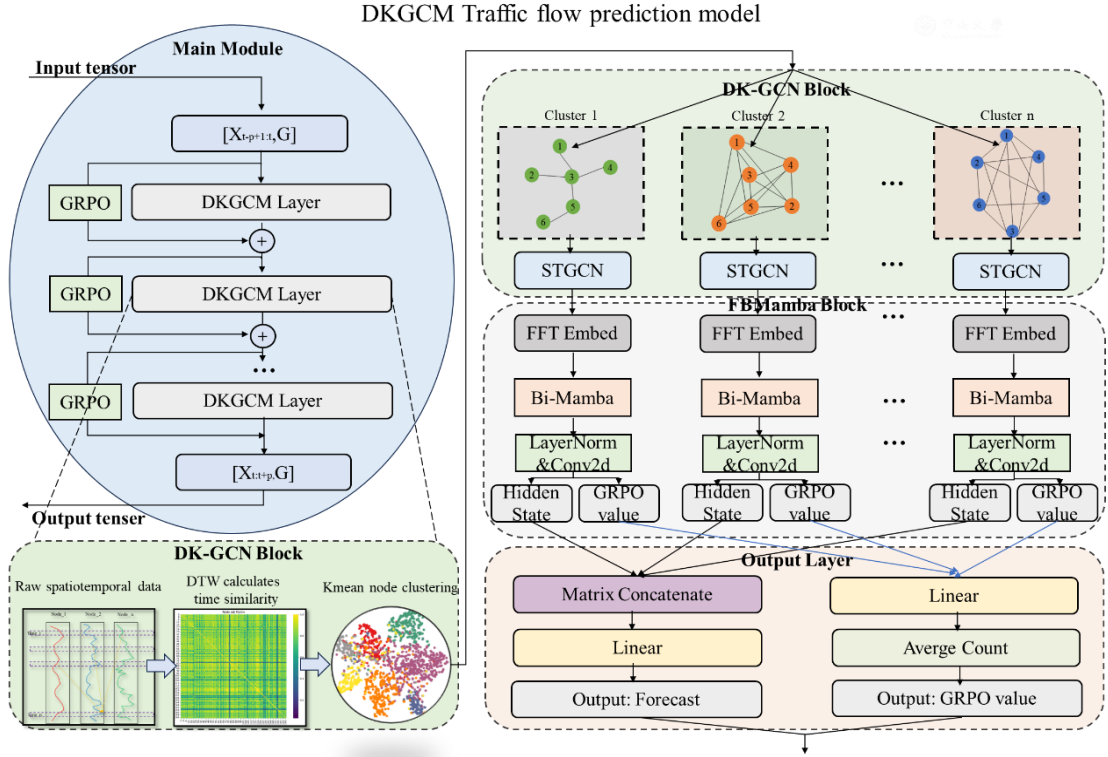

Figure 2 Structure of DKGCM traffic flow prediction model

For detailed algorithm descriptions of the spatial extraction module (DK-GCN), temporal extraction module (FBMamba), and loss optimization module (GRPO), please refer to Sections 4.1-4.3.

## 4.1 DTW-Kmeans-GCN(DK-GCN)

GCN, known for its ability to aggregate local topological information, is extensively used in graph-structured data modeling to capture spatial dependencies among adjacent nodes. However, traditional GCNs based on pre-defined graphs focus solely on spatial correlations and overlook temporal similarities between nodes. To address this limitation, we introduce Similarity-based Graphs that incorporate both spatial and temporal dynamics. The core of Similarity-based Graphs lies in the calculation of similarity. DTW is a robust similarity measurement technique for time series, capable of non-linearly aligning sequences to account for phase shifts while preserving underlying pattern similarities, offering clear advantages over traditional Euclidean distance metrics. In addition, it is necessary to cluster the nodes that share the same temporal patterns.

Therefore, Kmeans, as an efficient and cost-effective unsupervised clustering algorithm, which is less susceptible to issues such as empty clusters and performs well in partitioning data based on inherent structural patterns, is used to solve this problem. By integrating these three components, we propose a novel clustering-based graph convolution structure, DK-GCN, which enhances traditional predefined graph construction methods by dynamically capturing both temporal similarities and spatial correlations.

In summary, the DK-GCN temporal similarity clustering graph neural network consists of three main components: the DTW algorithm, the K-means clustering algorithm, and the GCN graph convolutional neural network. The functional structure of the algorithm is illustrated in Figure 3.

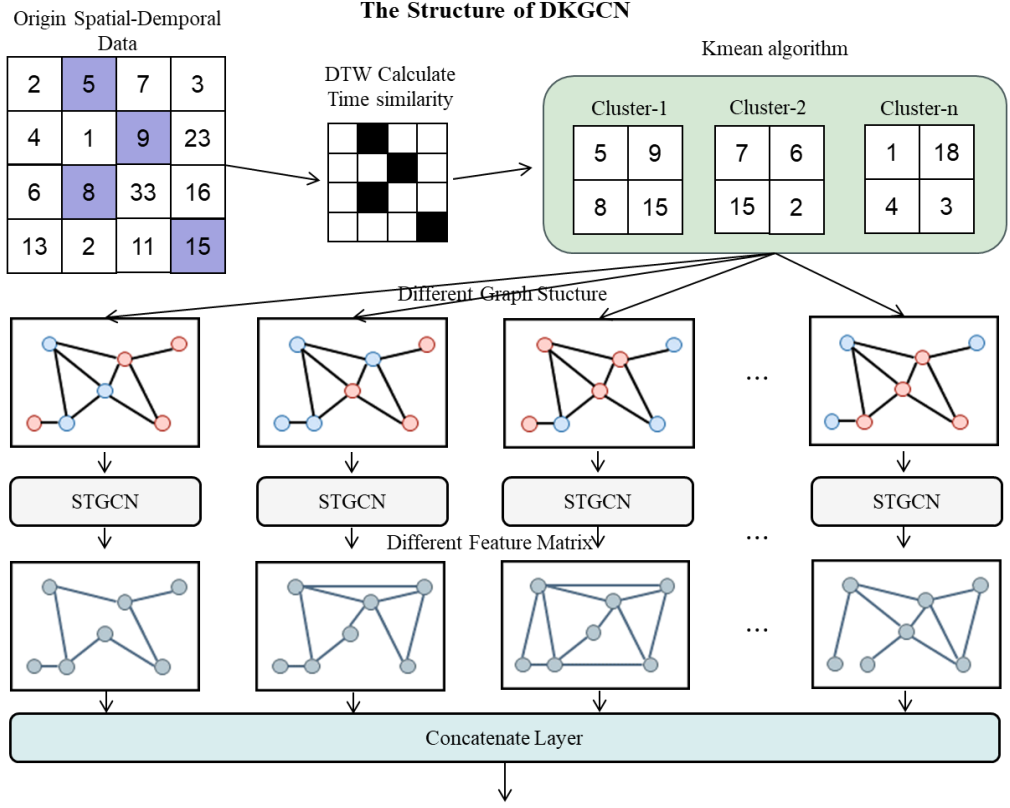

Figure 3 Functional composition of DK-GCN module

DTW is used to measure the similarity between two time series. It primarily aligns the time axes using dynamic programming and extracts the nonlinear features of the sequence over time. Given the traffic flow time series $X^a = [x_1^a, x_2^a, \cdots, x_t^a] \in \mathbb{R}^{t \times N}$ and $X^b = [x_1^b, x_2^b, \cdots, x_t^{ab}] \in \mathbb{R}^{t \times N}$ of two nodes, the Euclidean distance matrix between the two time steps $i\ and\ j$ is calculated as $D(i,j) =| x_i^a - x_j^b |$, Additionally, we define the cumulative distance matrix $M(i,j)$ to represent the minimum path sum from $(1,1)$ to $(i,j)$, as shown in Formula 2.

$$M(i,j) = D(i,j) + min\{M(i-1,j), M(i-1,j-1), M(i,j-1)\} \qquad (2)$$

Where *min*( ) represents the function of finding the minimum value. At the same time, the boundary conditions that must be met for the calculation of the cumulative distance matrix are shown in Formula 3.

$$M(1,1) = D(1,1), M(i,1) = \sum_{k=1}^{i} D(k,1), M(1,j) = \sum_{k=1}^{j} D(1,k) \qquad (3)$$

The DTW algorithm allows us to calculate the temporal feature similarity between different traffic nodes, followed by node clustering using the unsupervised K-means algorithm. The data is iteratively divided into *K* clusters, minimizing the squared distance between each data point and its cluster center.

In the K-means clustering algorithm, we begin by randomly selecting $K$ initial cluster centers $\{\mu_1, \mu_2, ..., \mu_k\}$, and assign each data point $x_i$ to the nearest cluster center $C_i$, as shown in Formula 4.

$$C_i = \arg \min_k \|x_i - \mu_k\|^2 \tag{4}$$

Next, recalculate the cluster center by taking the mean $\mu_k$, of the data points in the cluster, as shown in Formula 5.

$$\mu_k = \frac{1}{|C_k|} \sum_{x_i \in C_k} x_i \tag{5}$$

Finally, the assignment and update steps are repeated until the change in the cluster center is smaller than the threshold, or the maximum number of iterations is reached. Using the node clustering based on time feature similarity, we divide nodes with similar time series into the same category and input them into the GCN layer for spatial feature extraction. At each time step, we use graph convolution to process the signal and capture the spatial correlation of traffic information. The graph topology feature is represented by the Laplacian matrix, as shown in Formula 6.

$$\begin{aligned} L &= D - A \\ L_{\text{sym}} &= D^{-\frac{1}{2}} L D^{-\frac{1}{2}} = I - D^{-\frac{1}{2}} A D^{-\frac{1}{2}} \end{aligned} \tag{6}$$

Here, $D \in \mathbb{R}^{N \times N}$ represents the degree matrix, and $A \in \mathbb{R}^{N \times N}$ represents the adjacency matrix. To ensure the stability of the convolution operation, we apply regularization to convert the Laplacian matrix into a positive definite matrix, $L_{\text{sym}}$, and then perform Chebyshev polynomial recursion, as shown in Formula 7.

$$T_k(x) = 2x T_{k-1}(x) - T_{k-2}(x) \tag{7}$$

Where $T_0(x) = 1$ and $T_1(x) = x$. Consequently, the characteristic matrix $H^{(l)}$ of the traffic flow $X^{(l)}$ for each node in the road network, after being filtered by the kernel function $g_\theta$, is shown in Formula 8.

$$H^{(l)} = g_\theta * Gconv(X) = \sum_{k=0}^{K-1} \theta_k T_k(L_{\text{sym}}) X^{(l)} \tag{8}$$

Here, $Gconv()$ represents a convolution operation equivalent to the Chebyshev polynomial approximation. The convolution kernel $g_\theta$ extracts spatial feature weights from the 0th to ($K$-1)th neighboring nodes of each node, and an activation function layer outputs the result, as shown in Formula 9.

$$\tilde{A} = A + I, \quad \tilde{D} = \sum_j \tilde{A}_{ij}$$
$$H^{(l+1)} = \sigma\left(\tilde{D}^{-\frac{1}{2}}\tilde{A}\tilde{D}^{-\frac{1}{2}}H^{(l)}W^{(l)}\right) \tag{9}$$

Where $\tilde{A}$ is the matrix after adding self-loops, $\tilde{D}$ is the degree matrix of $\tilde{A}$, $H^{(l)}$ and $H^{(l+1)}$ are the node feature matrices of the $l$th layer and $l$-1th layer respectively, $W^{(l)}$ is the weight matrix, and $\sigma(\ )$ is the ReLU activation function.

## 4.2 Fourier Bidirectional Mamba(FBMamba)

At the time-series modeling level, time-related data exhibits dual characteristics across both the time domain and frequency domain. Time-domain features capture local signal variation patterns, while frequency-domain features reveal the global periodic structures. Traditional methods often struggle to capture these complementary aspects simultaneously. To address this challenge, we propose the FFTEmbedding mechanism, which decomposes time series signals into amplitude and phase spectra using the Fast Fourier Transform (FFT). The model then adaptively fuses time-domain and frequency-domain features, leveraging the global insight of the frequency domain alongside the fine-grained, contextual understanding of the time domain. This synergy enables the generation of richer and more robust sequence representations.

To address the challenges of efficiency and bidirectional dependency in long-sequence modeling, we draw inspiration from the BiTCN framework (D. Zhang et al., 2024), which has demonstrated strong performance in capturing temporal features within spatiotemporal prediction tasks. Building on this idea, we introduce a bidirectional Mamba structure (He et al., 2025), whose state-space model (SSM) efficiently captures long-term dependencies in both forward and backward directions while maintaining linear computational complexity. This bidirectional design significantly mitigates the information loss caused by the unidirectional nature of the original Mamba model. Consequently, we present the FBMamba layer, which integrates the Fast Fourier Transform with the bidirectional Mamba mechanism. The architectural design of this layer is illustrated in Figure 4.

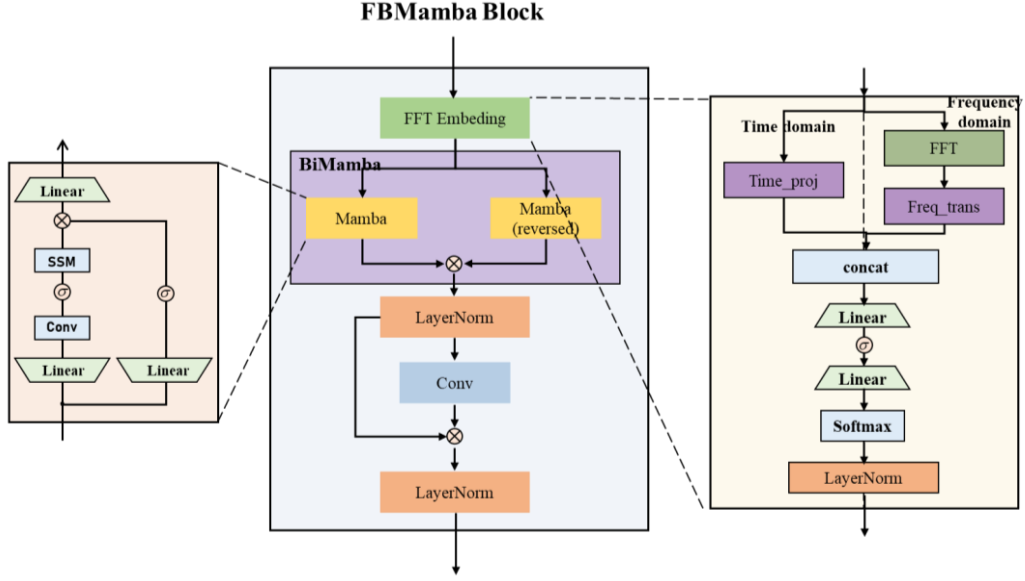

Figure 4 FBMamba module functional composition

As illustrated in Figure 4, the proposed FFTEmbedding layer performs a time-domain projection on the input node time series signal $\mathbf{X} \in \mathbb{R}^{B \times N \times L}$, where $B$ is the batch size, $N$ is the number of detectors, and $L$ represents the length of the time sequence, as defined in Formula 10.

$$\mathbf{T} = \text{Time\_proj}(\mathbf{X}) \in \mathbb{R}^{B \times N \times d_h} \tag{10}$$

Where Time_proj represents linear projection, which maps the time series length $L$ to dimension $d_h$. After completing the time domain projection, we perform frequency domain decomposition. First, we perform real-valued FFT calculation on each node time series, as shown in Formula 11.

$$\mathcal{F}(\mathbf{X}) = \text{rfft}(\mathbf{X}) \in \mathbb{C}^{B \times N \times K}, K = L/2 + 1 \tag{11}$$

Here, $\mathcal{F}(\mathbf{X})$ denotes the complex representation of the sequence $\mathbf{X}$ in the frequency domain layer $\mathbb{C}$, obtained via the real-valued Fast Fourier Transform (rfft). Since only the non-negative frequency components are retained, the resulting sequence length becomes $K = L/2 + 1$. We then decompose the complex-valued output into its amplitude spectrum $\mathbf{A}$ and phase spectrum $\mathbf{P}$, concatenate the two components, and apply a linear projection to map the concatenated representation back to the hidden dimension $d_h$, resulting in the vector $\mathbf{F}$, as defined in Formula 12.

$$\begin{aligned} \mathbf{A} &= |\mathcal{F}(\mathbf{X})|, \mathbf{P} = \angle \mathcal{F}(\mathbf{X}) \in \mathbb{R}^{B \times N \times K} \\ \mathbf{F} &= \text{Linear}\left([\mathbf{A}; \mathbf{P}]\right) \in \mathbb{R}^{B \times N \times d_h} \end{aligned} \tag{12}$$

We then adaptively fuse the time-domain feature vector $\mathbf{T}$ and the frequency-domain feature vector $\mathbf{F}$, as illustrated in Formula 13.

$$\mathbf{E} = \text{Linear}([\mathbf{T}; \mathbf{F}]) \in \mathbb{R}^{B \times N \times L}$$

$$\boldsymbol{\alpha} = \text{softmax}(\mathbf{E}) \in \mathbb{R}^{B \times N \times L}$$

$$\mathbf{Z} = \sum_{i=1}^{2} \alpha_i \cdot \mathbf{H}_i, \mathbf{H} = [\mathbf{T}, \mathbf{F}] \tag{13}$$

$$\mathbf{Z}' = \text{LayerNorm}(\text{Dropout}(\mathbf{Z}))$$

Here, $\mathbf{E}$ denotes the tensor obtained by concatenating the time-domain and frequency-domain feature vectors, followed by a linear projection, where $\mathbf{E} \in \mathbb{R}^{B \times N \times L}$. The scalar $\boldsymbol{\alpha}$ serves as a weighting factor to quantify the relative importance of the time-domain and frequency-domain features, enabling the adaptive selection of the fused feature $\mathbf{Z}$. Finally, a Dropout layer and a LayerNorm layer are applied to $\mathbf{Z}$ to produce the final time series representation $\mathbf{Z}'$ after Fourier-based processing.

After processing the traffic flow time series features using the Fourier transform, we address the causal limitation of the unidirectional Mamba by designing a bidirectional Mamba structure. This mechanism is then applied to extract temporal features more effectively, as described in Formula 14.

$$\mathbf{Y}_f = \text{Mamba}_{\text{forward}}(\mathbf{Z}')$$

$$\mathbf{Y}_r = \text{reverse}\left(\text{Mamba}_{\text{backward}}(\text{reverse}(\mathbf{Z}'))\right) \tag{14}$$

$$\mathbf{Y} = \mathbf{Y}_f + \mathbf{Y}_r \in \mathbb{R}^{B \times N \times d_h}$$

Here, $\mathbf{Y}_f$ denotes the time series features extracted by the forward Mamba mechanism, while $\mathbf{Y}_r$ represents those extracted by the backward Mamba mechanism. This bidirectional approach, similar to the BiTCN structure, enables more effective sequence modeling and produces the combined output $\mathbf{Y}$ after integrating information from both directions. Finally, a convolutional layer followed by a feedforward neural network performs feature scaling and produces the prediction output, as illustrated in Formula 15.

$$\mathbf{X}_1 = \text{Conv}_2\left(\sigma\left(\text{Conv}_1(\text{LayerNorm}(\mathbf{Y})^\top)\right)^\top\right), \sigma = \text{GeLU}$$

$$\mathbf{Z}_{\text{out}} = \text{LayerNorm}(\mathbf{X}_1) \tag{15}$$

We normalize the vector $\mathbf{Y}$ obtained from the bidirectional Mamba mechanism, then apply two convolutional layers, $\text{Conv}_1$ and $\text{Conv}_2$, for feature scaling. Finally, the scaled vector $\mathbf{X}_1$ is normalized to produce the final output tensor $\mathbf{Z}_{\text{out}}$.

## 4.3 GRPO

The training process of deep learning models is often based on static loss functions, such as MAE and MSE, serving as fixed optimization targets, while overlooking the effectiveness of dynamically adjusting loss weights and gradient propagation. Unlike traditional training paradigms, the introduction of reinforcement learning techniques can model the training process as a Markov Decision Process (MDP), thereby enabling dynamic adjustment of weights.

Reinforcement learning has evolved from Proximal Policy Optimization (PPO)(C. Yu et al., 2022) and Direct Policy Optimization (DPO) to Group Relative Policy Optimization (GRPO)(Ramesh et al., 2024), which plays a key role in Deepseek model today. However, the PPO algorithm relies on clipped targets（clipped）, which can lead to local oscillations or over-conservatism in the optimization process. In contrast, the DPO algorithm constrains the behavior strategy and the reference strategy through a KL divergence term, making it more suited for learning-based approaches. GRPO, on the other hand, is better suited for this optimization process, as it avoids the non-smooth objective function caused by clipping, resulting in a smoother and more stable optimization trajectory.

Integrating reinforcement learning strategies into deep learning models is a promising research direction for improving performance. In reinforcement learning, the strategy $\pi(a \mid s; \theta)$ defines the probability of taking action *a* in state *s*, with the parameter $\theta$, The objective of reinforcement learning is to maximize the expected cumulative reward $J(\theta)$, as shown in Formula 16.

$$J(\theta) = \mathbb{E}\left[\sum_{t=0}^{T} \gamma^t R_t\right] \quad (16)$$

Where $R_t$ represents the reward at time *t*, and $\gamma \in [0,1]$ is the discount factor that determines the importance of future rewards. This concept can be applied to the loss update strategy in spatiotemporal prediction models. Therefore, this paper integrates the GRPO algorithm to enhance the training module of the spatiotemporal prediction model.

GRPO evaluates the new policy primarily by comparing its output with that of the old policy, as shown in Formula 17.

$$L(\theta) = min(r_t A_t, \text{clip}(r_t, 1 - \epsilon, 1 + \epsilon) A_t) \quad (17)$$

$$r_t = \frac{\pi_{new}(a_t \mid s_t; \theta)}{\pi_{old}(a_t \mid s_t)}$$

Here, $r_t$ is the ratio between the new and old policies, $a_t$ is the action at time $t$, $s_t$ is the state at time $t$, and $A_t$ represents the advantage of action $a_t$ compared to the average. The clip( ) function constrains the value of $r_t$, preventing extreme updates. This mechanism ensures that if the new policy deviates too much from the old one, the update is constrained, thus preventing large changes that could lead to training instability. Inspired by the GRPO algorithm, we incorporate an advantage signal to adjust prediction results and enhance the loss function, as shown in Formula 18.

$$L_{\text{policy}} = -\mathbb{E}[min(r_t A_t, r_{\text{clipped}} A_t)]$$
$$r_{\text{clipped}} = \text{clip}(r_t, 1 - \epsilon, 1 + \epsilon) \quad (18)$$
$$Loss = MAE(\ ) + \tau L_{\text{policy}}$$

Here, MAE denotes the mean absolute error between predicted and true values, $L_{\text{policy}}$ is the policy loss (penalty function), and $r_{\text{clipped}}$ is the clipped ratio. This method enables a more stable training process and more accurate prediction results.

## 5. Experiments

### 5.1 Experiment Setting

We conducted a series of experiments using different datasets and time steps, focusing on three widely used California highway traffic datasets: PEMS04, PEMS07, and PEMS08. These datasets were selected due to their availability and extensive usage in traffic prediction research. The statistical details of these three public datasets are provided in Table 2.

Table 2 Statistical information of the datasets

| Model | Timesteps | Nodes | Time Range |
| --- | --- | --- | --- |
| PEMS04 | 16992 | 307 | 1/1/2018–2/28/2018 |
| PEMS07 | 28224 | 883 | 5/1/2017–8/31/2017 |
| PEMS08 | 17856 | 170 | 7/1/2017–8/31/2017 |

Based on the research of previous scholars, common metrics for measuring model performance include the Root Mean Square Error (RMSE) and Mean Absolute Error (MAE). Thus, we also adopted these two metrics in this study. The formulas for these evaluation metrics are provided below Equation (19).

$$\text{RMSE} = \sqrt{\frac{1}{n}\sum_{i=1}^{n}(\hat{X}_i - X_i)^2}$$

$$\text{MAE} = \frac{1}{n}\sum_{i=1}^{n}|\hat{X}_i - X_i|$$

(19)

To evaluate the performance of the proposed method, the following spatiotemporal traffic flow prediction methods from the existing literature are selected as benchmarks.

- DCRNN: This model combines diffusion convolution with a recurrent neural network, simulates the spatial diffusion process via bidirectional random walk, and employs gated recurrent units (GRU) for time series data processing.
- STGCN: This model utilizes a Graph Convolutional Network to capture spatial dependencies and combines a 1D convolution to model temporal dependencies.
- GWNet: This model replaces traditional graph convolution with the graph wavelet transform, enabling more efficient processing of the spatial dimension in graph data.
- AGCRN: This model learns the graph structure by introducing an adaptive mechanism to better handle dynamically changing graph-structured data.
- GMAN: This model effectively captures complex spatiotemporal correlations through spatiotemporal embedding and a multi-head attention mechanism.
- MTGNN: This graph neural network model addresses multivariate time series prediction by constructing a graph structure between time series to predict future values.
- STID: This model achieves accurate spatiotemporal data prediction through spatiotemporal data embedding and identity attachment combined with a simple neural network structure.
- STAEFormer: This model integrates both temporal and spatial embeddings, along with temporal and spatial axial attention mechanisms, to enable effective multivariate prediction.

- **STD-MAE**: This model enhances spatiotemporal prediction performance by integrating spatiotemporal patch embedding with a spatiotemporal dual-mask attention mechanism.

Additionally, the following multivariate prediction methods are also adopted for comparisons.

- **Informer**: This model introduces the ProbSparse self-attention mechanism within the original Transformer architecture, significantly reducing the time and memory complexity of self-attention computations.
- **Dlinear(Zeng et al., 2023)**: This model adopts a decomposition strategy to split time series data into multiple linear components, followed by sequential linear transformations for prediction.
- **iTransformer**: This model employs an inverted Transformer structure specifically optimized for time series forecasting.
- **MGCN**: This model addresses long-sequence prediction tasks by integrating a Graph Convolutional Network with a bidirectional Mamba structure.

For the experiments, we uniformly split the three public datasets into training, validation and test sets, using an 8:1:1 ratio. Model parameters were standardized by setting the batch size to 32, the number of epochs to 100, and ensuring a consistent input length across all models. We employed L1 Loss as the loss function and used the Adam optimizer with a learning rate of 0.001 for model training. Baseline model hyperparameters were configured based on the default settings provided in the publicly available code. Experiments were conducted on a system equipped with an Intel i7-8700 CPU and an NVIDIA RTX 3090 GPU, and the DKGCM model's codes are available at https://github.com/dragonlsq/DKGCM-Traffic-prediction-model/tree/main.

## 5.2 Results

The statistics of the model experiment results, along with the compared model calculations, are presented in Table 3 below.

Table 3 Evaluations of our model and baselines

| Model | | PEMS04 | | PEMS07 | | PEMS08 | |
|---|---|---|---|---|---|---|---|
| Method | Horizen | @6 | @12 | @6 | @12 | @6 | @12 |
| Dlinear | MAE | 31.7 | 38.2 | 27.4 | 38.6 | 27.2 | 32.6 |
|  | RMSE | 47.3 | 55.9 | 40.2 | 55.1 | 39.9 | 47.0 |
| Informer | MAE | 24.7 | 25.6 | 32.4 | 33.8 | 26.1 | 29.2 |
|  | RMSE | 39.3 | 40.5 | 56.0 | 57.4 | 40.9 | 46.4 |
| iTransformer | MAE | 20.3 | 23.3 | 21.5 | 24.8 | 14.9 | 18.0 |
|  | RMSE | 33.3 | 36.8 | 34.8 | 38.1 | 23.8 | 28.7 |
| DCRNN | MAE | 19.7 | 21.7 | 21.2 | 24.1 | 15.2 | 17.7 |
|  | RMSE | 31.4 | 34.2 | 34.4 | 38.9 | 24.3 | 27.1 |
| STGCN | MAE | 19.6 | 21.1 | 21.7 | 24.2 | 15.9 | 17.6 |
|  | RMSE | 31.3 | 33.5 | 35.4 | 39.5 | 25.3 | 28.0 |
| GWNet | MAE | 18.9 | 20.5 | 20.3 | 22.8 | 14.7 | 16.2 |
|  | RMSE | 30.3 | 32.5 | 33.4 | 37.1 | 23.5 | 25.9 |
| AGCRN | MAE | 19.5 | 20.6 | 20.7 | 22.7 | 15.7 | 17.5 |
|  | RMSE | 31.5 | 33.5 | 34.5 | 37.9 | 25.0 | 27.9 |
| GMAN | MAE | 18.8 | 20.0 | 20.3 | 22.3 | 14.6 | 15.7 |
|  | RMSE | 30.9 | 31.3 | 33.3 | 36.4 | 24.1 | 26.5 |
| MTGNN | MAE | 19.5 | 21.0 | 20.8 | 23.6 | 15.3 | 16.8 |
|  | RMSE | 32.0 | 34.7 | 33.9 | 38.1 | 24.4 | 27.0 |
| MGCN | MAE | 18.5 | 19.7 | 19.9 | 21.8 | 14.4 | 15.4 |
|  | RMSE | 30.3 | 31.8 | 33.5 | 35.1 | 24.2 | 24.8 |
| STID | MAE | 18.3 | 19.6 | 19.6 | 21.5 | 14.2 | 15.6 |
|  | RMSE | 29.9 | 31.8 | 32.9 | 36.3 | 23.6 | 25.9 |
| STAEFormer | MAE | 18.2 | 19.5 | 19.2 | 20.8 | 13.5 | 14.2 |
|  | RMSE | 30.2 | 31.4 | 32.6 | 34.2 | 23.3 | 24.3 |
| STD-MAE | MAE | 17.8† | 19.5† | **18.7** | 20.1† | 13.5† | 14.1† |
|  | RMSE | 29.3† | 31.3† | 31.4† | 32.9† | 22.5† | 23.8† |
| Ours | MAE | **17.58** | **19.41** | 19.07† | **20.02** | **13.46** | **13.99** |
|  | RMSE | **29.26** | **30.62** | **29.86** | **31.47** | **20.08** | **21.03** |

(In the Table, **Bolding** denotes optimal solutions and † denotes suboptimal solutions)

As shown in Table 3, classic spatiotemporal graph-based prediction models (e.g., DCRNN and STGCN) outperform many state-of-the-art (SOTA) multivariate prediction models (e.g., iTransformer). This indicates that incorporating graph structures significantly enhances spatiotemporal prediction, while multivariate models often overlook variable correlations, leading to decreased prediction performance.

Additionally, from the perspective of performance metrics, our proposed DKGCM model demonstrates an average reduction of 5.21% in MAE and 13.46% in RMSE compared to the STID baseline model, which performs well in spatiotemporal prediction tasks. These results highlight the strong performance of the DKGCM model across tasks with varying time steps.

Furthermore, the model enables the generation of prediction results for each spatial subregion. Specifically, we present the prediction outcomes across the three datasets-PEMS04, PEMS07, and PEMS08-under the settings of clustering parameter $N=5$ and prediction step size of 12, as shown in Table 4 and Figure 5.

Table 4 Prediction index results of each sub-region after clustering

| DataSet | | Cluster1 | Cluster2 | Cluster3 | Cluster4 | Cluster5 | SUM |
|---|---|---|---|---|---|---|---|
| PEMS08 | MAE | 21.36 | 10.38 | 17.30 | 15.04 | 5.85 | 13.99 |
| (horizon=12) | RMSE | 32.08 | 14.75 | 25.45 | 23.48 | 9.45 | 21.03 |
| PEMS07 | MAE | 14.74 | 32.64 | 6.94 | 20.32 | 25.46 | 20.02 |
| (horizon=12) | RMSE | 25.82 | 48.46 | 13.18 | 33.33 | 36.58 | 31.47 |
| PEMS04 | MAE | 14.15 | 18.12 | 20.27 | 34.20 | 10.33 | 19.41 |
| (horizon=12) | RMSE | 21.87 | 29.07 | 32.91 | 53.96 | 15.28 | 30.62 |

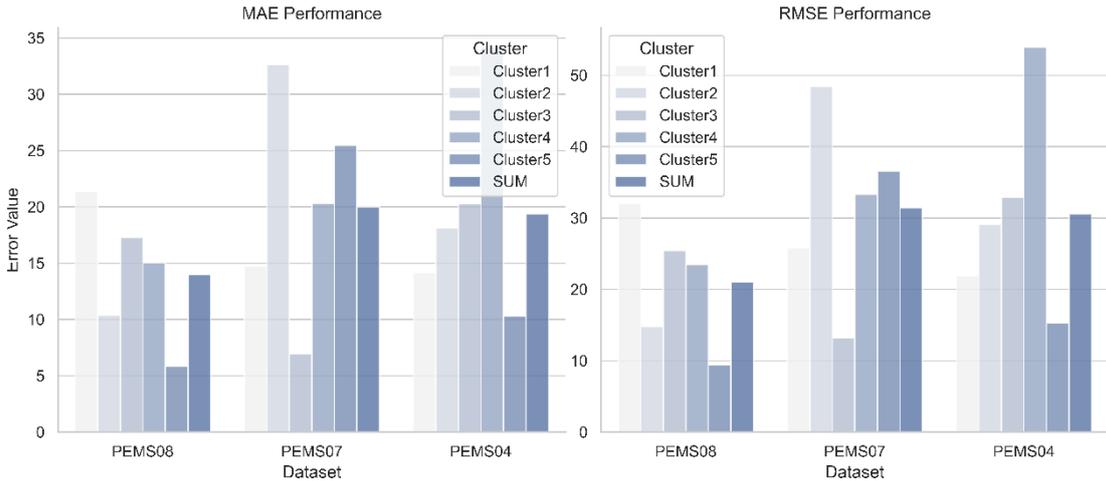

Figure 5 Comparison of prediction effects of each spatial sub-region after clustering

Figure 5 illustrates that time series clustering divides the entire spatial area into sub-regions with varying traffic flow magnitudes, resulting in distinct prediction performances across different sub-regions. For instance, Cluster 5 in PEMS04 and PEMS08, as well as Cluster 3 in PEMS07, exhibit relatively low traffic flow values, which correspond to lower absolute error and root mean square error. By segmenting the space into homogeneous sub-regions, nodes within each sub-region share similar temporal patterns and amplitudes. This spatial coherence facilitates the subsequent Fourier-based Bidirectional Mamba in capturing stable temporal trends, thereby improving the overall prediction accuracy. These findings also highlight the contribution of the DK-GCN module in enhancing model performance by leveraging meaningful spatiotemporal structure.

## 5.3 Discussion

We visualized the node adjacency matrix, constructed using Laplace feature mapping in the traditional GCN method, and the node temporal feature cosine similarity correlation matrix in our proposed DK-GCN method, as shown in Figure 6. There are notable differences between the node temporal features and the adjacency features of the traditional spectral method. This confirms that the traditional predefined graph method exhibits feature deviations, supporting the effectiveness of our proposed node temporal clustering method.

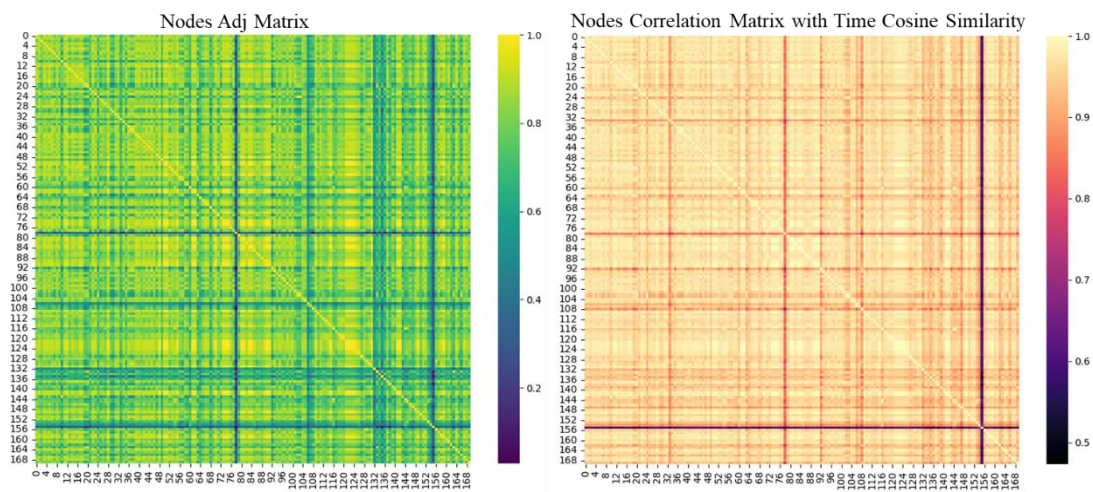

Figure 6 Node adjacency feature matrix and node temporal feature cosine similarity feature matrix

Therefore, we abandoned the traditional approach of constructing a predefined graph using adjacency features and opted for a graph construction method based on node temporal clustering. To further examine the spatiotemporal feature relationships of various node types after temporal clustering, we applied the commonly used t-SNE dimensionality reduction method to illustrate the spatiotemporal heterogeneity of these node categories after DTW and K-means clustering, as shown in Figure 7.

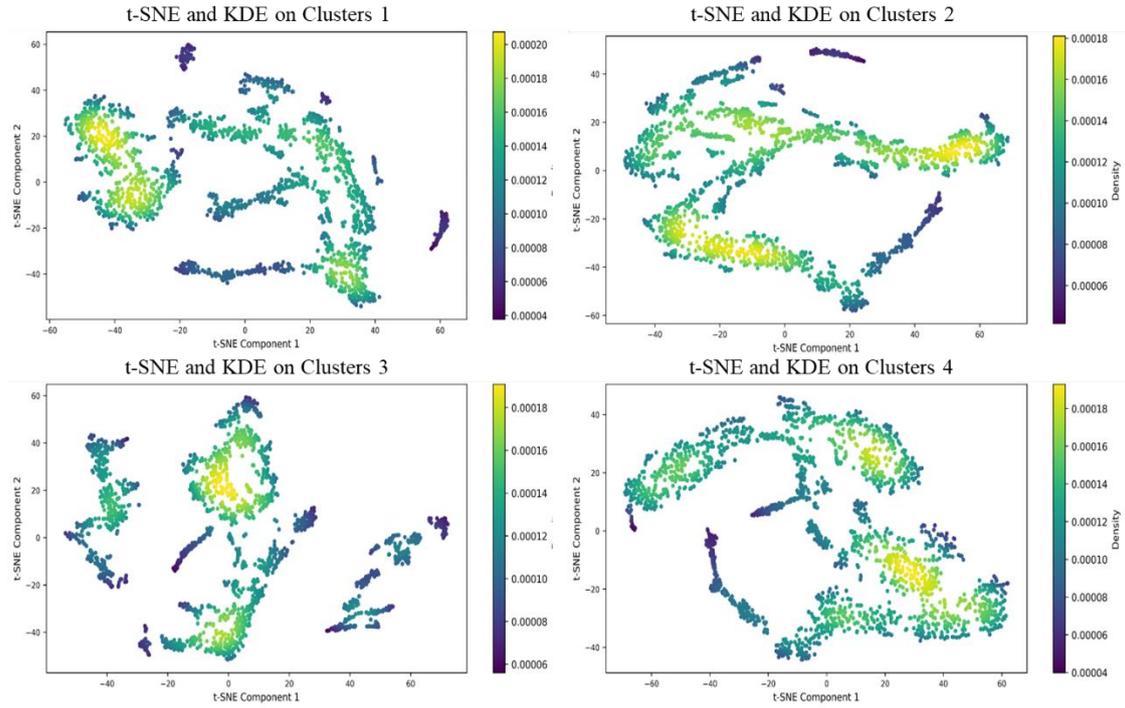

Figure 7 Spatial heterogeneity between node clusters of different categories

After clustering, we observe that nodes of different categories exhibit distinct spatiotemporal distribution characteristics, resulting in significant spatial heterogeneity. Therefore, the spatial clustering method in DK-GCN helps separate nodes with similar spatiotemporal characteristics and allows for the joint prediction of these nodes, thereby enhancing the model's prediction performance.

### 5.3.1 Ablation Study

To verify the effectiveness of each module in the proposed DKGCM model, we conducted the following ablation experiments: (1) RemDKGCN: removes the clustering graph convolution module; (2) RemFBMamba: removes the FBMamba module; (3) RemGRPO: removes the GRPO module; (4) RepGRU: replaces the FBMamba module with the GRU module; (5) RepTrans: replaces the FBMamba module with the Transformer Encoder module (nhead=4, encoder_layers=3). the ablation experiment settings are shown in Table 5.

Table 5 The module settings of Ablation experiment   (✓: used, ✗: unused)

| Model | DK-GCN | FBMamba | GRPO | GRU | Transformer |
|---|---|---|---|---|---|
| DKGCM | ✓ | ✓ | ✓ | ✗ | ✗ |
| RemFBMamba | ✓ | ✗ | ✓ | ✗ | ✗ |
| RemDKGCN | ✗ | ✓ | ✓ | ✗ | ✗ |
| RemGRPO | ✓ | ✓ | ✗ | ✗ | ✗ |
| RepGRU | ✓ | ✗ | ✓ | ✓ | ✗ |
| RepTrans | ✓ | ✗ | ✓ | ✗ | ✓ |

The ablation experiments were conducted on the PEMS08 dataset, comparing the MAE and RMSE results for the experimental control group under different time step scenarios (horizon=6, 12, 24, and 48), as shown in Figure 8.

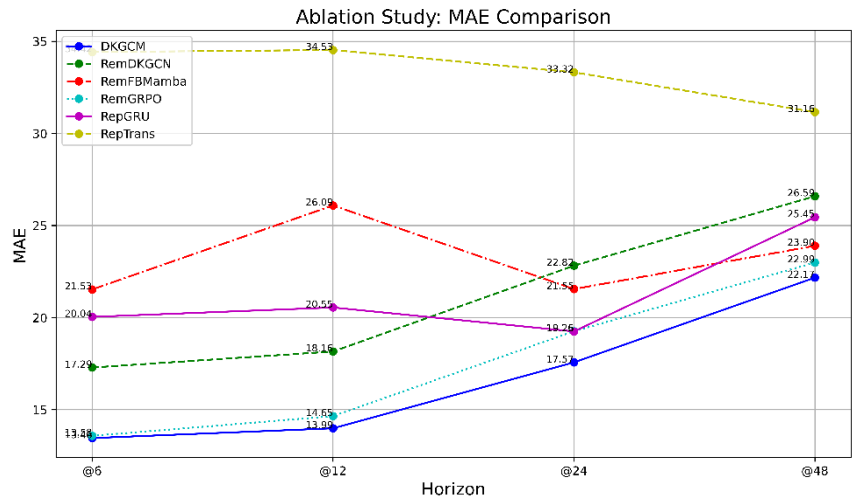

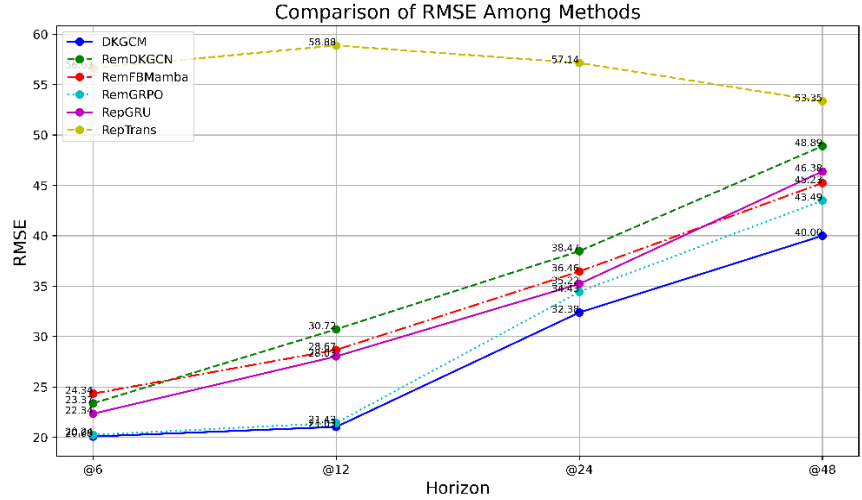

Figure 8 Ablation experiment results of PEMS08 dataset

The ablation experiment for RemFBMamba shows that its accuracy on the PEMS08 dataset is significantly lower than that of DKGCM, indicating that the integrated FBMamba module significantly improves prediction accuracy. Compared to DKGCM, RepGRU and RepTrans show lower accuracy, demonstrating that FBMamba has certain advantages over GRU-based and Transformer-based models in extracting temporal features of traffic flow. Similarly, the prediction accuracy of the model after RemDKGCN is significantly reduced, indicating that the spatial feature extraction method considering node clustering can further improve prediction accuracy. The performance of the module after RemGRPO is slightly reduced, suggesting that the GRPO reinforcement learning fusion LOSS optimization strategy has some effect on improving prediction performance, though it is not significant. Therefore, the performance of DKGCM outperforms other variants, confirming the effectiveness of each component in the proposed traffic flow spatiotemporal prediction framework.

In addition, the experimental results for different time steps show that shorter time steps (@6,@12) yield better prediction performance, while the prediction effect gradually weakens as the time step increases. This also reveals the trend of short-term versus long-term prediction. In contrast, the prediction error of RepTrans with the

Transformer module gradually decreases as the number of time steps increases, highlighting the Transformer model's advantage in capturing long-term dependencies. Due to the Attention mechanism that enables the calculation of global feature similarities, Transformer-based models are better at capturing global dependencies. This observation further illustrates both the strengths and limitations of the Transformer structure.

### 5.3.2 Sensitivity Analysis

The ablation experiment shows that both the DK-GCN and FBMamba modules are key components of the model. We further performed a sensitivity analysis on three parameters during model training. These parameters are the number of clusters ($N$), sequence length (Seq_Len), and the number of hidden layers (Hidden_dim) in the DKGCM model. The experiment was conducted on the PEMS08 dataset, and the results are shown in Figures 9, 12, and 13.

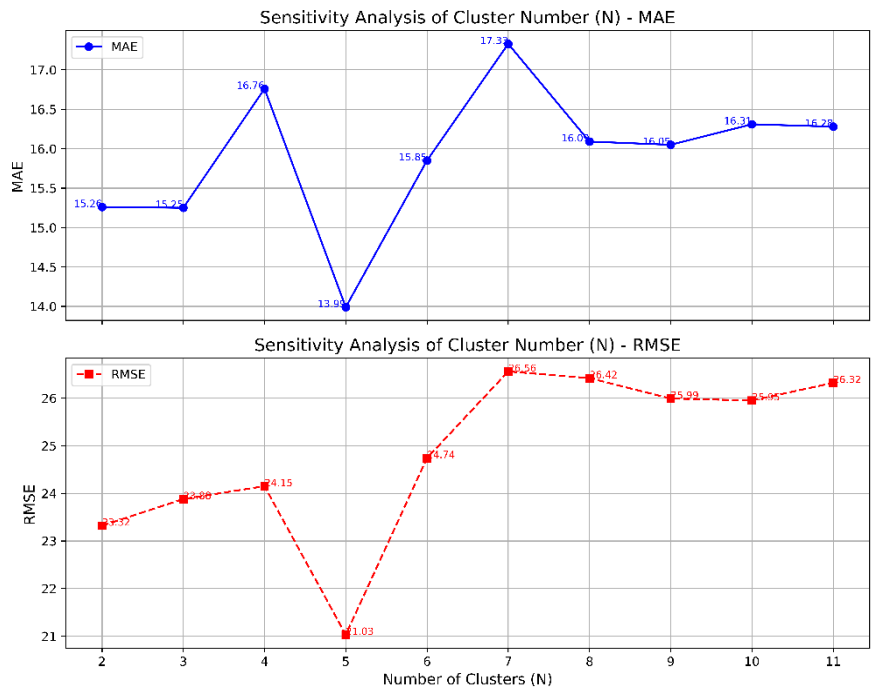

Figure 9 Sensitivity analysis results of the number of clusters N in the PEMS08 dataset

To observe the impact of the parameter $N$ in the clustering graph convolution model, we set the sequence length (Seq_Len) to 12 and the number of hidden layers (Hidden_dim) to 64. From Figure 9, we see that the model's prediction performance is optimal when the number of clusters is set to 5. Settings with too few classifications (e.g., $N$=2,3) or too many classifications (e.g., $N$=7,8) reduce prediction accuracy, which is largely due to the time series characteristics of different nodes. When the parameter is too low, the model classifies nodes with large time series differences into the same category, reducing accuracy. When the parameter is too high, the improvement in classification is minimal, and the model's complexity increases significantly. We further illustrate the node clustering effect in Figure 10.

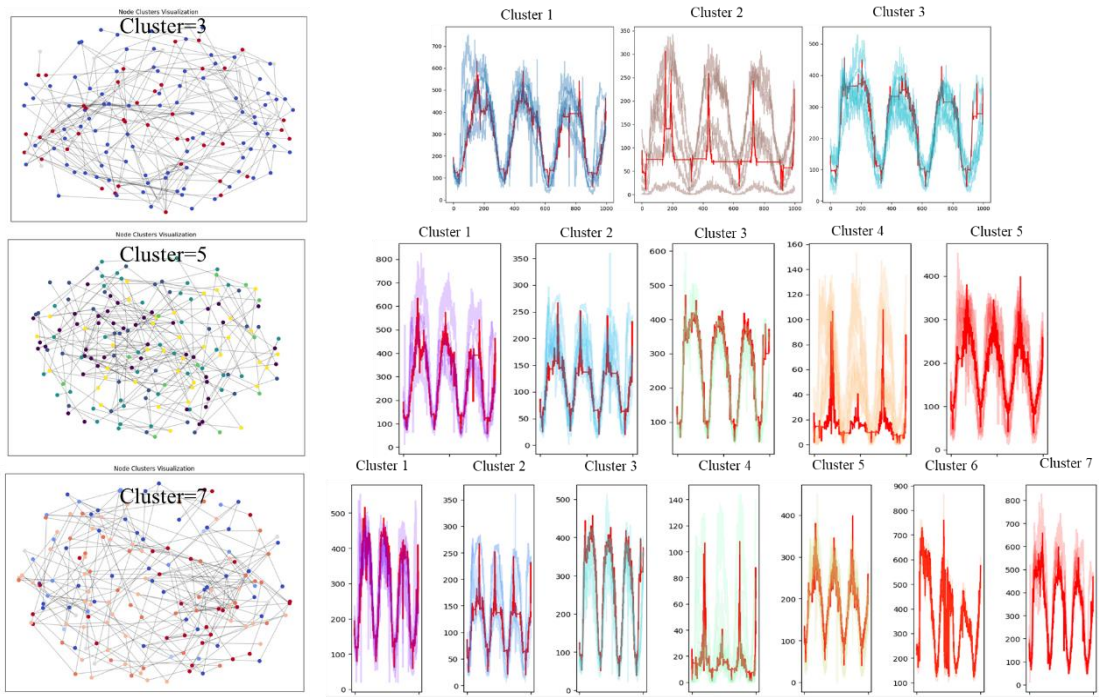

Figure 10 Time series distribution of nodes in different clusters

As shown in Figure 10, a low parameter setting ($N$=3) tends to classify nodes with significant time series differences into the same cluster, such as Cluster 2, whereas a moderate setting ($N$=5) rarely leads to this issue. Furthermore, a high setting ($N$=7) can also group dissimilar nodes into the same cluster, as seen in Cluster 4. These excessive differences increase prediction error, resulting in poorer performance compared to the moderate setting ($N$=5). This finding supports the results of our sensitivity analysis on the number of clusters ($N$).

To validate the effectiveness of the clustering graph convolution module in the DK-GCN model under real-world conditions and to assess the appropriateness of the chosen number of clusters ($N$), we replaced the K-means clustering algorithm with the K-Shape clustering algorithm in the DK-GCN module. The resulting time series clustering of each node is shown in Figure 11.

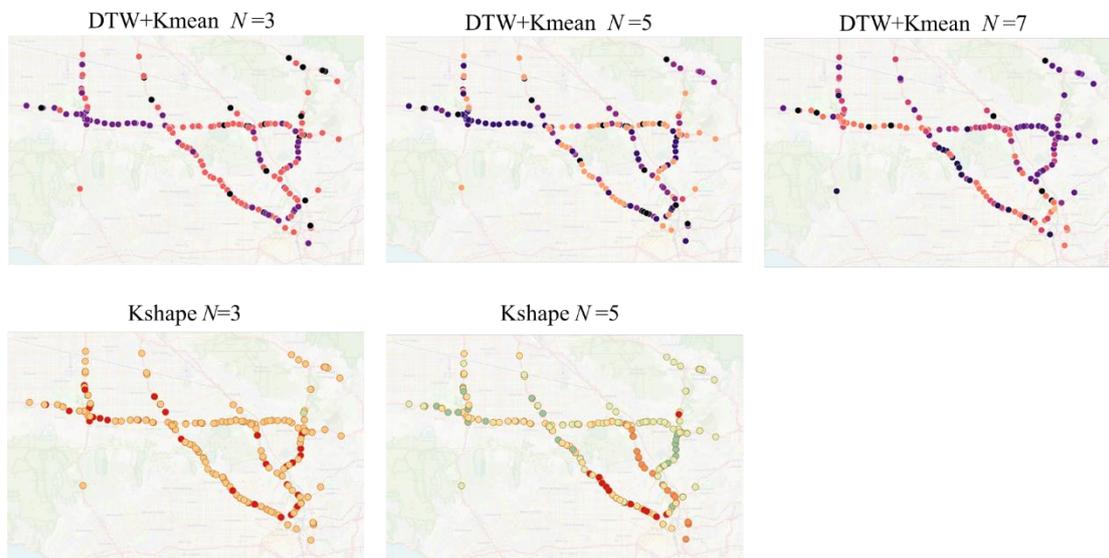

Figure 11 Spatial distribution of traffic nodes with different cluster number settings

Figure 11 shows that the node distribution resulting from the clustering algorithm differs significantly from the traditional distribution based on a predefined graph. Similar to how DCRNN simulates traffic dynamics through a diffusion process, the DK-GCN's time series-based node clustering better reflects real-world traffic flow patterns. We found that setting the number of clusters ($N$) to 5 yields a node classification that aligns well with observed traffic patterns, such as continuous flow, intermittent flow, and congestion. However, setting $N$ too low or too high results in overly simplistic or overly complex classifications that do not match real-world traffic conditions, further supporting our conclusion. We also conducted a preliminary comparison with other time series clustering algorithms, such as K-Shape, and found that their performance in realistic scenarios was inferior to DTW and K-means. As shown in Figure 11, traffic flow distinctions become less clear, and excessively high values of N (e.g., N = 6 or 7) often result in empty clusters.

In addition to evaluating the impact of the parameter (Seq_Len), we conducted experiments with the number of clusters ($N$) set to 5 and the number of hidden units (Hidden_dim) set to 64. The experimental results are presented in Figure 12.

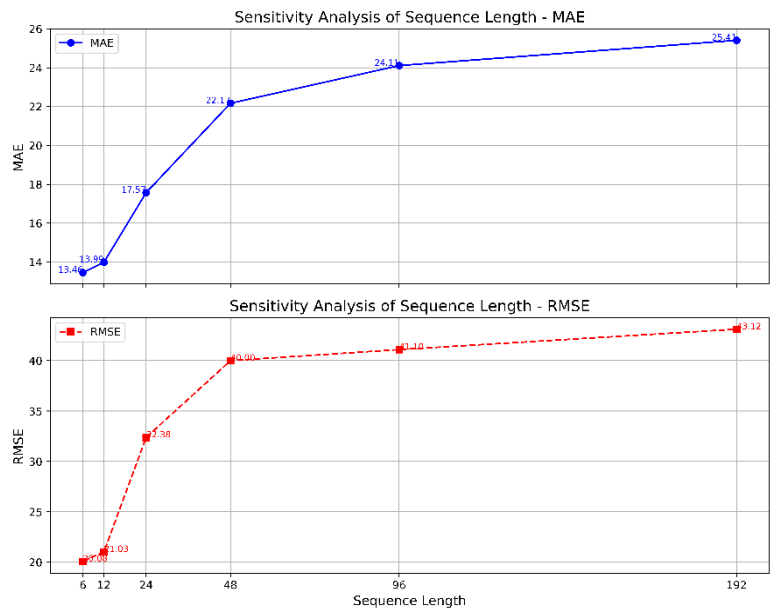

Figure 12 Sensitivity analysis results of the sequence length Seq_Len of the PEMS08 dataset

Figure 12 shows a clear pattern regarding the sequence length (Seq_Len): shorter sequences lead to lower prediction errors, while longer sequences result in higher errors. However, the rate of increase in error slows as the sequence length grows. This observation aligns with the intuitive expectation that short-term traffic flow predictions are more accurate than long-term forecasts. Additionally, we performed a sensitivity analysis on the number of hidden units (Hidden_dim) in the model, as illustrated in Figure 13.

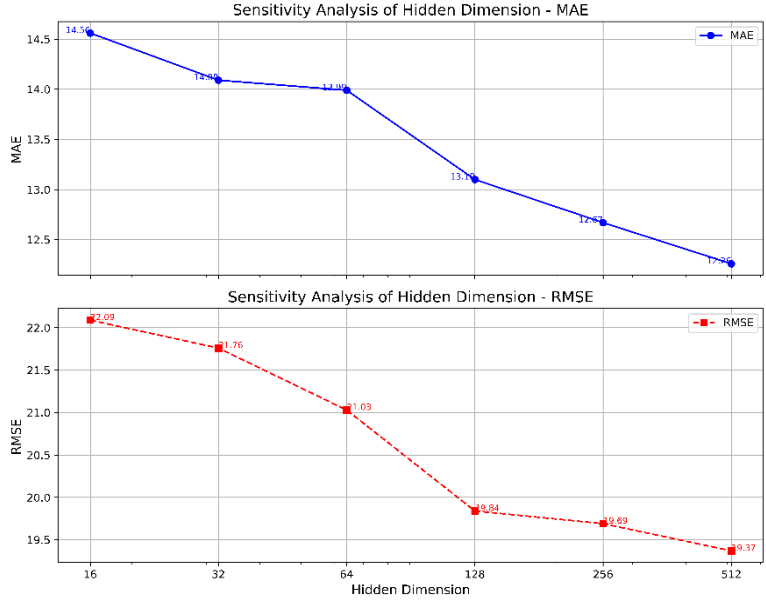

Figure 13 Sensitivity analysis results of the number of Hidden_dim in the PEMS08 dataset

As shown in Figure 13, when the number of clusters (*N*) is fixed at 5 and the sequence length (Seq_Len) is set to 12, increasing the number of hidden units improves the model's predictive performance, with a significant improvement observed in the range of Hidden_dim = 16–128. However, as Hidden_dim increases further to the range of 256–512, the improvement becomes marginal and even begins to decline. This suggests that higher-dimensional hidden layers can better capture complex temporal relationships in diverse traffic patterns. However, excessively large settings may lead to increased computational complexity and longer training times. Therefore, a moderate hidden dimension, such as Hidden_dim = 512, provides the best predictive performance for the model.

In addition, to verify the effectiveness of the dynamic loss function incorporating the reinforcement learning-based GRPO strategy improvement, we conducted a sensitivity analysis on the key parameter $r_t$, which controls the gradient update (the clipping ratio $r_t$ was set to 0.1, 0.2, and 0.3, respectively). A control experiment was also conducted using the RemGRPO experimental group, in which the GRPO module was removed. We performed the experiments on the PEMS08 dataset, and the loss curves of the training set under different parameter settings are presented in Figure 14.

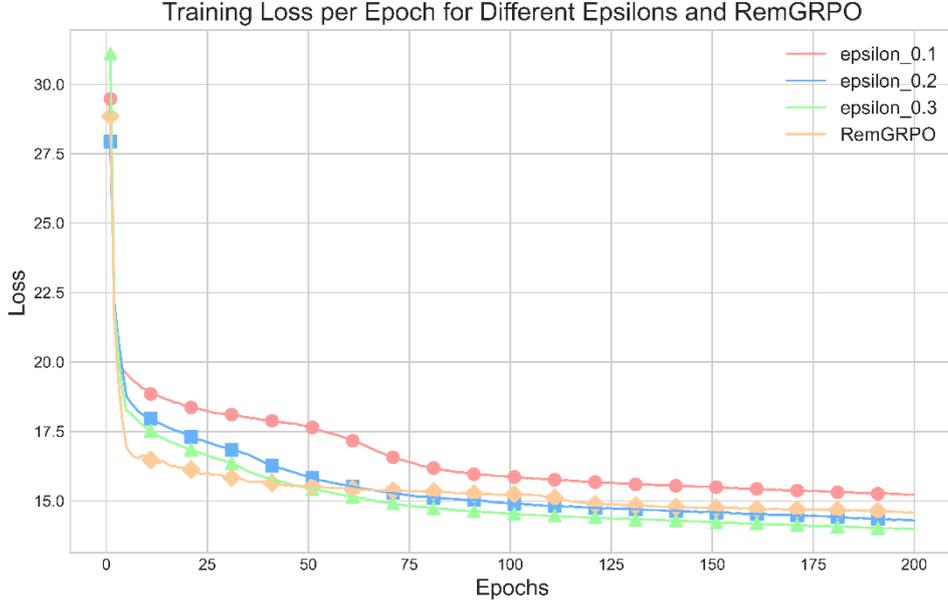

Figure 14 Training Loss per Epoch for Different Epsilons and RemGRPO in the PEMS08 dataset

From Figure 14, we observe that the prediction model incorporating the GRPO strategy to optimize the loss function achieves better prediction performance than the static loss function model when the $r_t$ parameter is set to 0.2 and 0.3, while it performs poorly when $r_t$ is set to 0.1. Furthermore, since $r_t$ serves to limit the update magnitude of the reinforcement learning strategy, it can be seen that the loss decline rate of the GRPO-optimized models is slower compared to the RemGRPO model. This indicates that although the GRPO models converge more slowly, they achieve better final prediction performance. A smaller $r_t$ value leads to slower updates, which is a limitation of the method, sacrificing some training speed in exchange for slight performance improvements. Based on the sensitivity analysis, setting $r_t$ to 0.3 yields the best predictive performance.

## 6. Conclusion

This paper proposes a novel spatiotemporal traffic flow prediction model,

DKGCM. The model first introduces a temporal clustering graph convolutional network (DK-GCN) to extract spatial features, then incorporates the FBMamba model to capture temporal dependencies, and finally applies the GRPO reinforcement learning strategy to optimize training. The proposed model maintains low computational complexity while effectively handling complex data dependencies, significantly enhancing prediction accuracy. Extensive experiments on three public datasets demonstrate the model's accuracy in traffic flow prediction, and ablation studies further validate the effectiveness of each component.

However, this study has several limitations. Although the introduction of reinforcement learning algorithms enhances the training process and potentially improves prediction accuracy, setting an excessively low clipping ratio may lead to longer training times. In addition, when facing a larger data set scenario, the time spent on DTW and K-means clustering will also be longer.

Future research should focus on enhancing the model's memory efficiency to ensure reliable performance when processing large-scale datasets. For instance, more efficient clustering algorithms could be explored to minimize memory usage. Additionally, the signal modal decomposition algorithm could be integrated into the Mamba time series feature extraction module to reduce sequence fluctuations and enhance data stability.